\documentclass[10pt,twocolumn,letterpaper]{article}

\usepackage{icb}
\usepackage{times}
\usepackage{epsfig}
\usepackage{mathptmx} 
\usepackage{graphicx}
\usepackage{amsmath}
\usepackage{amssymb}
\usepackage{algorithm}
\usepackage{algpseudocode}
\usepackage{color}

\input{latex.macro}

\icbfinalcopy 

\ificbfinal\fi
\begin{document}

\title{Fast Matching by 2 Lines of Code for Large Scale Face Recognition Systems}

\author{Dong Yi, Zhen Lei, Yang Hu and Stan Z. Li\thanks{Stan Z. Li is the corresponding author.} \\
Center for Biometrics and Security Research \&
National Laboratory of Pattern Recognition \\
Institute of Automation, Chinese Academy of Sciences, Beijing, China\\
dyi, zlei, yhu, szli@cbsr.ia.ac.cn\\
}

\maketitle
\thispagestyle{empty}

\begin{abstract}
In this paper, we propose a method to apply the popular cascade classifier into
face recognition to improve the computational efficiency while keeping high
recognition rate. In large scale face recognition systems, because the probability of
feature templates coming from different subjects is very high, most of the matching
pairs will be rejected by the early stages of the cascade. Therefore, the cascade can
improve the matching speed significantly. On the other hand, using the nested structure
of the cascade, we could drop some stages at the end of feature to reduce the memory
and bandwidth usage in some resources intensive system while not sacrificing the
performance too much. The cascade is learned by two steps. Firstly, some kind
of prepared features are grouped into several nested stages. And then, the threshold
of each stage is learned to achieve user defined verification rate (VR). In the paper, we take a landmark
based Gabor+LDA face recognition system as baseline to illustrate the process
and advantages of the proposed method. However, the use of this method is very generic and
not limited in face recognition, which can be easily generalized to other biometrics as a post-processing module.
Experiments on the FERET database show the good performance of our baseline and an experiment on
a self-collected large scale database illustrates that the cascade can improve the matching speed significantly.
\end{abstract}

\section{Introduction}
\label{sec:intro}

Recently, face recognition technologies are applied in more and more applications
with big data, such as, face recognition for large scale social network services~\cite{face-com},
and face recognition system for surveillance. These large scale systems usually contain
many millions of templates in the database, which are deployed centralized or distributed.
And they need to deal with the requests from the users or other image sources (\eg Surveillance
Camera) continuously. Big data and high intensive request bring about many strict requirements to the
efficiency of face recognition systems. The efficiency of the systems are mainly determined
by two factors: the size of feature template (Storage and Transmission) and the speed of
template matching (Computation). In this paper, we will focus on these two factors and
propose a generic and lightweight method for efficient large scale face recognition.

The computational complexity of the most popular matching algorithms are linear with respect
to the number of templates $n$ and the dimension of feature $d$, such as Euclidean distance,
Cosine similarity and so on. There are two kinds of methods to reduce the computation:
approximated nearest neighborhood methods~\cite{Yan-IJCB-2011} and partial feature based filtering~\cite{Wu-PAMI-2011}.
The details of these methods will be described in the next section.
In this paper, mainly inspired by~\cite{Guo-ICCVW-2001}, we propose a novel fast matching method by applying the cascade
classifier~\cite{Viola-IJCV-04} from face detection to face recognition. For face recognition,
especially in large scale applications, the matching between the probe and the gallery set is
an asymmetry classification problem, in which the majority of matching pairs are negative (two images with
different identity). This is the case that the cascade classifier can work efficiently.

While feature template has been grouped into several stages, we can drop several stages at the end of the cascade to obtain smaller feature template. Small template can obviously save the storage space and improve the transmission speed of system, but it will also reduce the recognition rate. Therefore, we need to consider the trade-off between feature length and performance for specific needs. The spirit is very similar to the Scalable Video Coding (SVC)~\cite{Bennett-MILCOM-2008}
in H.264 video compression standard. SVC standardizes the encoding of a high-quality
video bitstream that also contains one or more subset bitstreams. A subset video bitstream
is derived by dropping packets from the larger video to reduce the bandwidth required for the
subset bitstream. The subset bitstream can represent a lower spatial resolution, lower
temporal resolution, or lower quality video signal. Similarly, the cascaded feature template
contains many subset feature bits, and the subset has lower recognition rate.

In summary, the cascade can provide two advantages for large scale face recognition systems: (1)
quickly reject negative face pairs to improve the matching speed; (2) according to the specific
storage and computation requirements, supply a scalable structure for user to select the size of feature.
To illustrate the advantages, we propose a EBGM-like~\cite{Wiskott-PAMI-97}
baseline system. First, a face detector and ASM are used to localize serval facial landmarks. Then,
multi-scale and orientation Gabor~\cite{Lades-Gabor-93} features are extracted on the landmarks.
Finally, LDA is applied to enhance the discrimination and reduce the dimensionality of feature. The
similarity of features are evaluated by Cosine metric.

The main contributions of the paper are as follows:
\begin{enumerate}
\item By studying the asymmetric property of large scale face recognition problem, we apply cascade classifier
    for fast face matching. As a generic and lightweight post-processing module, the cascade classifier can be used widely to improve the speed of other biometric systems;
\item We propose a new concept ``feature subset'' for biometrics by cutting-off several stages
of the cascade, and give the relationship between feature length and recognition rate by experiments;
\item To illustrate the advantages of our method, we propose an EBGM-like~\cite{Wiskott-PAMI-97} baseline face recognition  method, the performance of which is comparable to many state-of-the-art methods~\cite{Tan-TIP-2010, Xie-TIP-2010, Hussain-BMVC-2012}.
\end{enumerate}


\section{Related Work}

Fast algorithm for large scale pattern recognition problems has a long history in the field of image retrieval,
but don't get much attention in face recognition community. The most popular fast algorithm for large scale
image retrieval is approximated nearest neighborhood, such as k-d tree~\cite{friedman-an-algorithm-toms-77} and
Locality-Sensitive Hashing (LSH)~\cite{Datar-SCG-2004, Bennett-MILCOM-2008}. But \cite{Wu-PAMI-2011} has found
these approximated nearest neighbor search methods do not work well with high-dimensional face features, that means
their performance degrade quickly in face recognition with database increasing.

In \cite{Wu-PAMI-2011}, Wu \etal proposed a multi-reference re-ranking approach for large scale face recognition.
The main idea is originated from the query expansion techniques in text information retrieval. Firstly, many local
features of face components are used to get a small candidate set from the large gallery. Then a binary global
feature is used to re-rank the candidate set to get the final result. Experiments show the performance of their
algorithm is comparable with the linear scan system using the state-of-the-art face feature. On a database containing
one million face images, the speed-up ratio is about 8x comparing to the linear scan system. \cite{Liu-ECCV-2008} also
used similar way to improve the speed of face recognition by sifting the gallery according to rank. Our method is
closely related to these two methods, the idea is using partial or simple features to reject the irrelevant
samples as quickly as possible.

Inspired by the ideas of popular Hashing based methods~\cite{Datar-SCG-2004, Bennett-MILCOM-2008}, Yan \etal~\cite{Yan-IJCB-2011} proposed a Similarity Hashing (SH) for large scale face recognition, and got good performance on a database containing 100,000 face images. Although SH has achieved 30x speed-up ratio in the experiments, it is very memory consuming and need extra tens of Giga-Bytes to store the hash index for every samples in the gallery. Because our method exploits the asymmetric structure of the data in large scale face recognition, comparing to the above methods, our cascade structure not only can obtain high speed-up ratio but also has the lowest performance loss.

Cascade is a kind of extreme unbalanced tree to deal with asymmetric two-class problem. The most successful application of cascade is face detection~\cite{Viola-IJCV-04}, in which the cascade was used to reject non-face samples at each node (or stage) and nearly allow all face samples to pass. In \cite{Guo-ICCVW-2001} the cascade was used for face recognition, but its advantage was not noticed and analyzed in the context of large scale face recognition. In this paper, we take face matching as a two-class problem, then large face recognition problem is exactly an asymmetric classification problem. Given a probe, the majority of samples in gallery should have different identity with the probe. Therefore, we borrow the cascade structure from face detection to large scale face recognition in this case. We will see that the cascade can also work well for face recognition. The process of cascade learning in face recognition is simpler than that in face detection, because we only need to learn the threshold of each stage but the strong classifier.

\section{Baseline Face Recognition Method}
\label{sec:baseline}

To illustrate the concept and the advantages of the proposed method, we will discuss the
steps and algorithms base on a baseline face recognition method. The baseline is similar with the
popular EBGM and Bochum/USC system~\cite{Wiskott-PAMI-97, Okada-Bochum-98}. Firstly, we detect 76 facial landmarks
on a face image, and normalize the face image according to some reference points (\eg two eyes).
Then Difference of Gaussian (DoG) filter~\cite{Tan-TIP-2010} is used for illumination normalization. Second, we extract
multi-scale Gabor feature on the 76 landmarks. And then a holistic subspace learning is applied on
the Gabor features to enhance the discriminative ability. The detail steps are described in the
following subsections.

\subsection{Face Normalization}

Because face detection is relative irrelevant to this paper, we just skip this step. After face
detection, we localize the facial landmarks by Active Shape Model (ASM)~\cite{Cootes-ASM-95}.
ASM is composed by three parts: shape model, local experts and optimization strategy.

In the most ASM variants, shape model is usually PCA~\cite{Fukunaga-book-90}, and we follow this model. For local experts,
we use LBP~\cite{Ahonen-ECCV-04} feature and Boosting classifier for each landmark, which is similar with the method
in~\cite{Yi-IJCB-2011}. Based on the output of Boosting classifiers, we can get a confidence map
for each landmark. These confidence maps are feed to a Landmark Mean-Shift procedure~\cite{Saragih-IJCV-2011}.
Then we can get the final positions of all facial landmarks. For robustness and efficiency,
the optimization process is repeated several times on two scales.

\begin{figure}
\centering
\includegraphics[width=0.183\textwidth]{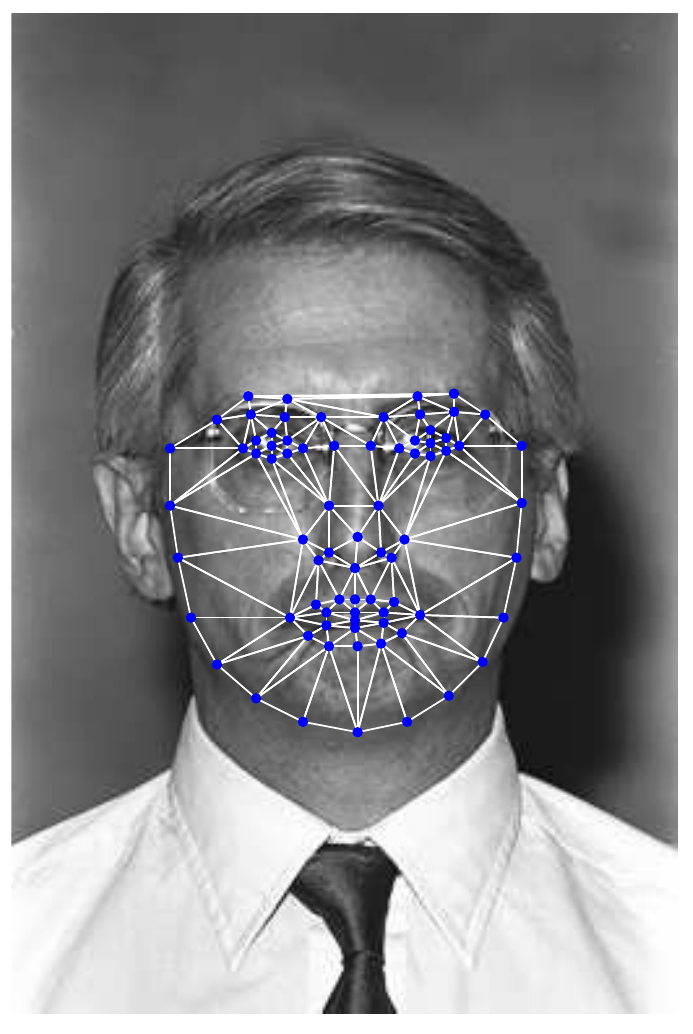}
\includegraphics[width=0.18\textwidth]{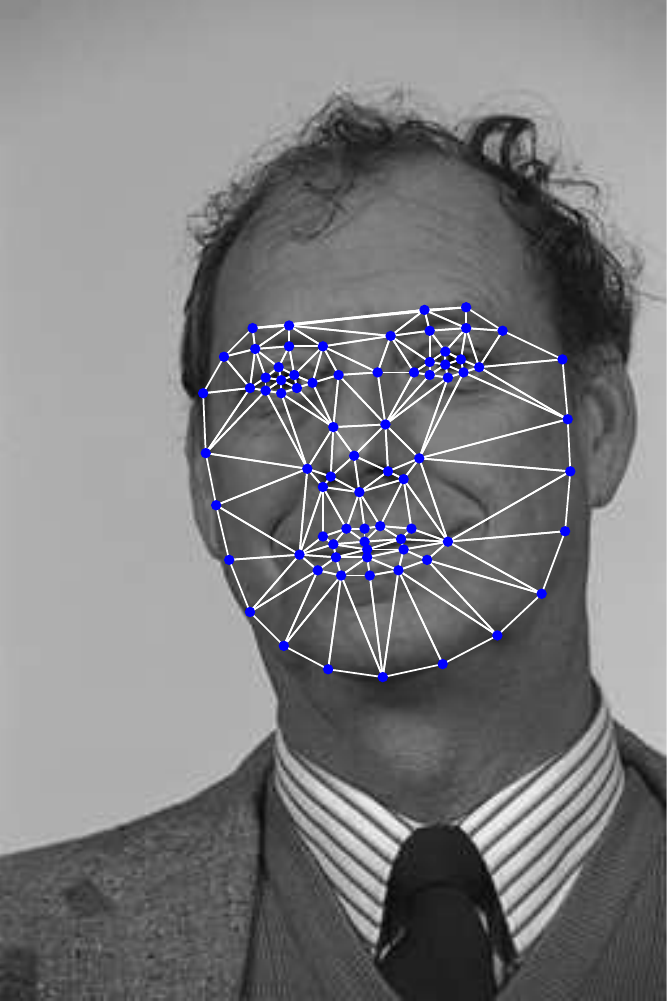}
\caption{Two face images in FERET database and their 76 facial landmarks localized by our ASM landmarker.}
\label{fig:76pts}
\end{figure}

\begin{figure}
\centering
\includegraphics[width=0.36\textwidth]{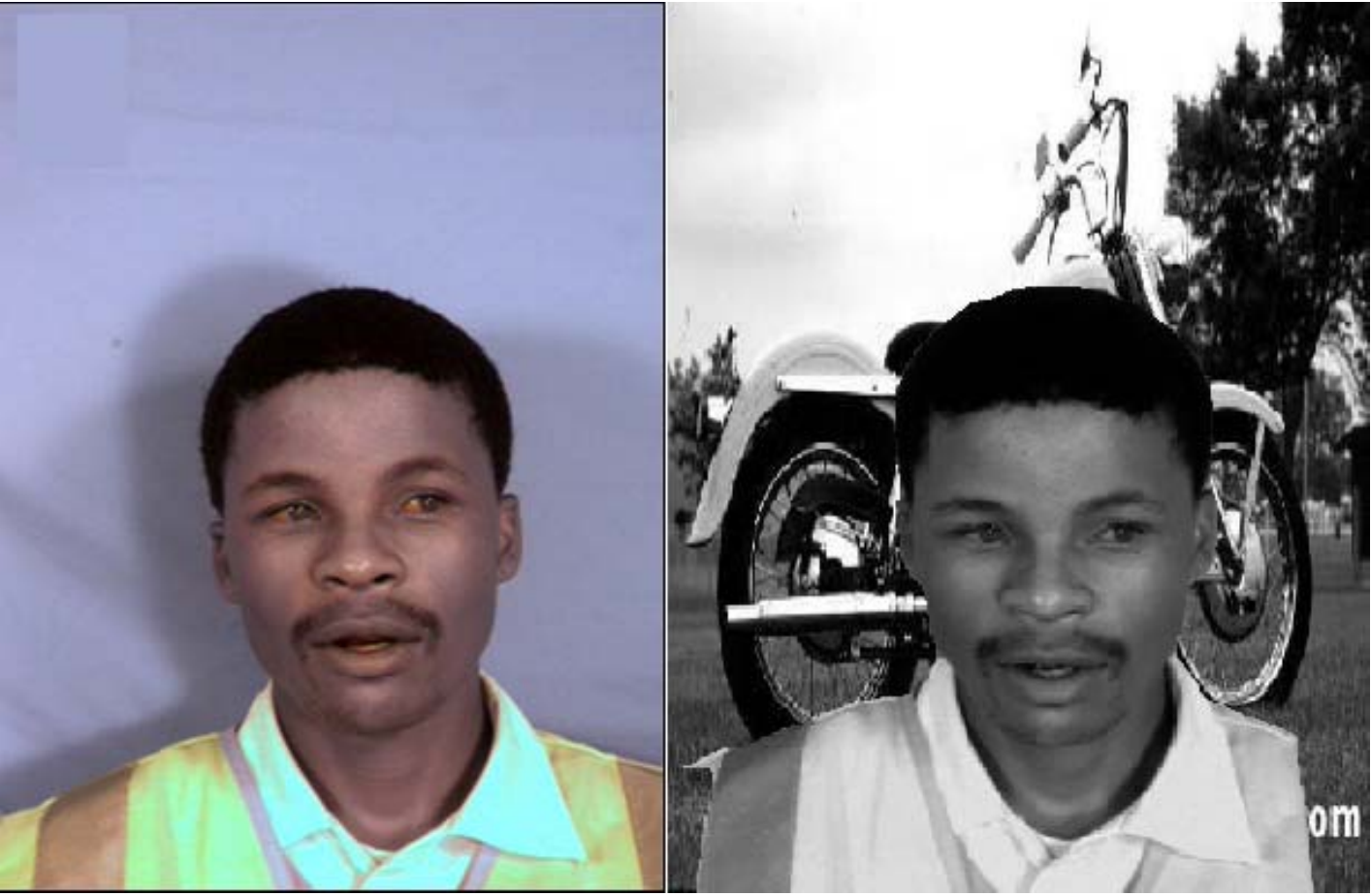}
\caption{Sample images in the MUCT training set for the ASM landmarker. Left: A color face image in the MUCT database, which has uniform background. Right: A face image is converted to gray scale and the background is replaced by a random image from the Internet.}
\label{fig:muct}
\end{figure}

The training set of our landmarker is constructed from the MUCT database~\cite{Milborrow-PRASA-2010}.
Three views (a, d and e) with small pose variations are used for training. Because the background of
images in the MUCT are almost uniform, we replace them with some random backgrounds and mirror all
images to augment the dataset (see Figure~\ref{fig:muct}). The uniform background of the face images
are segmented by GrabCut~\cite{Rother-ACM-TG-2004}, which is initialized by the results of face detection.
Figure~\ref{fig:76pts} shows two example images in the FERET database~\cite{Phillips-FERET-PAMI-00}
and their 76 facial landmarks localized by the landmarker, from which we can see the landmarks are
robust to small pose variations and have good precision for the next steps.

\begin{figure}
\centering
\includegraphics[width=0.15\textwidth]{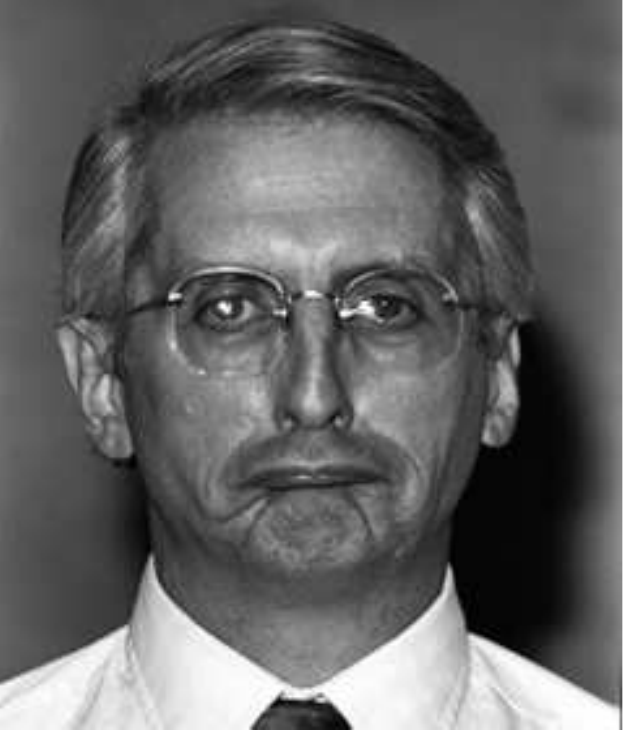}
\includegraphics[width=0.15\textwidth]{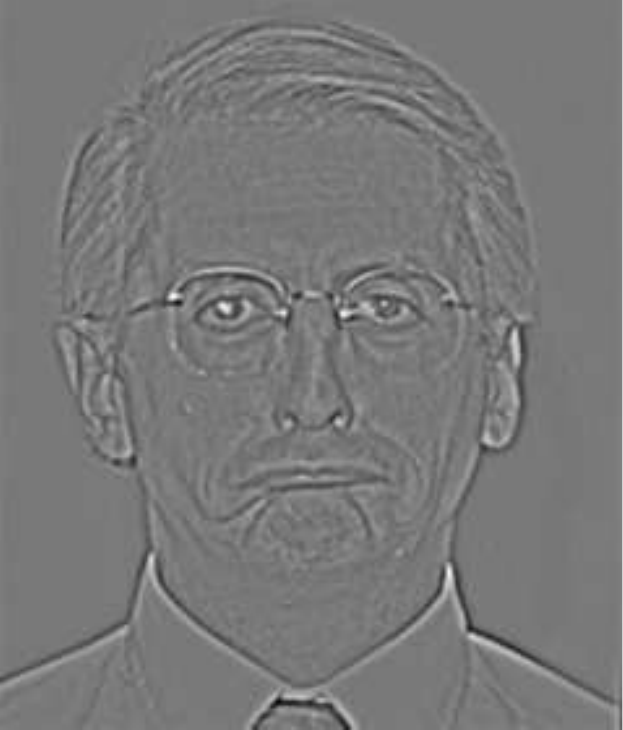}
\caption{An aligned face image and its corresponding DoG filtered image.}
\label{fig:DoG}
\end{figure}

For geometry normalization, two eyes in the 76 landmarks are selected as reference points. Based on
the reference points, the face images are normalized to a standard frame with eye distance $= 60$
pixels by a similarity transformation. Note that the above similarity transformation is also applied
on the 76 landmarks. And then DoG filter~\cite{Tan-TIP-2010} is used to
reduce the influence of illumination on the face images. The parameters of DoG are: $\gamma=0.2$,
$\sigma_0=0.8$, $\sigma_1=1.6$, $do\_norm=0$. Figure~\ref{fig:DoG} shows an aligned face image and
the DoG filtered image.

\subsection{Gabor Feature and LDA}

Given an aligned face image and the 76 landmarks, we extract local features on the landmarks by
a Gabor wavelet, which is described in~\cite{Okada-Bochum-98}.
\begin{equation}
\psi_{\bfk, \sigma}(\bfx) = \frac{k^2}{\sigma^2} e^{\frac{k^2}{-2\sigma^2} x^2} \{e^{i \bfk \bfx}-e^{-\frac{\sigma^2}{2}}\}
\end{equation}

\begin{figure}
\centering
\includegraphics[width=0.3\textwidth]{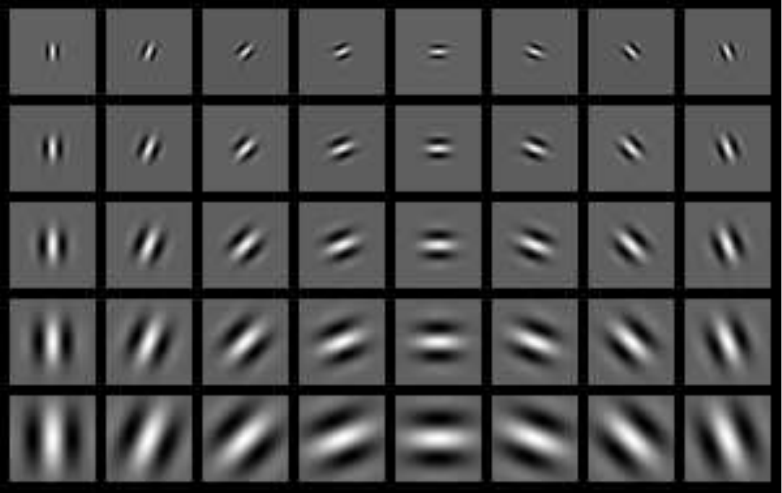}
\caption{The real part of the Gabor wavelet in 5 resolutions and 8 orientations.}
\label{fig:Gabor}
\end{figure}

The wavelet is a plane wave with wave vector $\bfk$, restricted by a Gaussian envelope,
the size of which relative to the wavelength is parameterized by $\sigma$. The second term
in the brace removes the DC component. Following the popular way, we sample the space of wave
vectors $\bfk$ and scale $\sigma$ in a discrete hierarchy of 5 resolutions (differing by half-octaves)
and 8 orientations at each resolution (See Figure~\ref{fig:Gabor}), thus giving $5\times8=40$ complex
values for each landmark. Because the phase information is sensitive to image shift or
mis-alignment, we just drop the phase and use the amplitude as feature for face recognition.

Merging the feature values at all landmarks together, we get a feature vector with $76\times40=3040$
dimensions. To reduce the dimensionality of feature and remove the redundant information, Linear
Discriminant Analysis (LDA)~\cite{Belhumeur-97} is used to learn a low dimensional subspace. In the LDA subspace, the
similarity of feature vectors are evaluated by Cosine metric.

\begin{equation}
s(\bfx, \bfy) = \frac{\bfx^T \bfy}{\sqrt{\bfx^T \bfx \bfy^T \bfy}}
\label{equ:cosine}
\end{equation}

In practice, we usually normalize the feature vector $\bfx$ and $\bfy$ to unit length as $\bfx'$ and
$\bfy'$. Then the Equation~(\ref{equ:cosine}) can be written as

\begin{equation}
s(\bfx, \bfy) = s(\bfx', \bfy') = \bfx'^T \bfy'.
\label{equ:dot_product}
\end{equation}

\section{Cascade Learning}
\label{sec:cascade}

As described in Section~\ref{sec:intro}, the objective of the cascade is to improve the speed of large
scale face recognition and supply a flexible way to get a trade-off between the feature length and
recognition rate.

How to construct the cascade is feature and matching algorithm dependent. In this paper, we construct
a specific cascade based on the Gabor-LDA feature and Cosine metric in the baseline method
(see Section~\ref{sec:baseline}).

\subsection{Structure}

\begin{figure}
\centering
\includegraphics[width=0.45\textwidth]{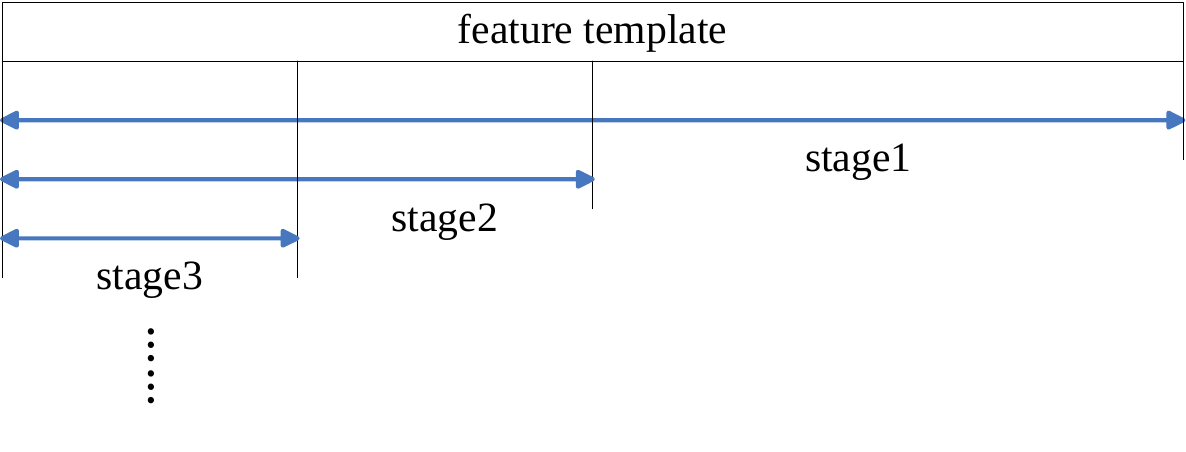}
\caption{The structure of the cascade classifier.}
\label{fig:SFT_structure}
\end{figure}

The cascade has been widely used in face detection since the famous work of Viola and Jones~\cite{Viola-IJCV-04}.
After that, there are many cascade variations arise for many specific applications, such as Nested cascade~\cite{Huang-ICPR-2004},
Soft cascade~\cite{Bourdev-CVPR-2005}, Boosting Chain~\cite{Xiao-ICCV-2003} and so on.
To use the existing features efficiently, we adopt the nested structure. As shown in Figure~\ref{fig:SFT_structure},
we divide the feature template into several nested stages from coarse to fine. The first level ``stage1'' is
actually the original feature template, which has the highest recognition rate. The ``stage2'' and ``stage3''
are coarse levels with lower recognition rate, and the feature template can be divided further.

When the structure of the cascade is determined, we can group the Gabor-LDA features in this way.
In the experiments of this paper, the size of feature is 428, and the feature template is divided
into a seven-stage structure. The sizes of the stages are: 6, 13, 26, 53, 107, 214, 428. For other applications,
the number of stage and the size of each stage can be chosen by experience or experiments. The
recognition rate of each stage will be reported in the experiments. In resource constrained
applications, we can drop some stages at the end of feature template to save the storage and
transmission overhead.


\subsection{Learning}


The target of the cascade classifier is to improve the feature matching speed. While training the
cascade, the multi-class samples are firstly converted to positive (sample pairs from same subject)
and negative samples (sample pairs from different subject) by cross matching. Given two samples $\bfx'$
and $\bfy'$, the training sample is constructed by Equation~(\ref{equ:construct}) as follows
\begin{equation}
\bfb = \bfx' \odot \bfy',
\label{equ:construct}
\end{equation}
where $\odot$ denote element-wise product.
Given the nested structure of the cascade and the training samples, we need to learn the threshold of each stage to
achieve some user defined Verification Rate (VR) (see Alg.~\ref{alg:cascade_learning}).
To reduce the false reject rate of positive samples, we usually set the VR of each stage nearly to 100\%. In this paper,
we set the verification rates for the seven stages as: $99.9\%$, $99.9\%^2 = 99.8\%$, $99.9\%^3 = 99.7\%$, $99.9\%^4 = 99.6\%$, $99.9\%^5 = 99.5\%$, $99.9\%^6 = 99.4\%$, $99.9\%^7 = 99.3\%$.

\begin{algorithm}
\caption{Cascade learning procedure.}
\label{alg:cascade_learning}
\begin{algorithmic}[1]
\State {\bf Input}: Positive samples $\{P_{ij}\}$, where $i = 1, 2, \cdots, d$, $j = 1, 2, \cdots, n$; $n$ is sample count; $d$ is feature dimension. Note that $\{P_{ij}\}$ are constructed from the training set by Equation~(\ref{equ:construct}). $\bfm$ is the size of each stage. $\bfv$ is the user defined VR of each stage. Stage count $sn= 7$.
\State {\bf Output}: Threshold $\bft$ of each stage.
    \State Let cumulative similarity $\bfs = \bf0$, $i = 0$, where the length of $\bfs$ is $n$.
    \For{$k = 0$ to $sn - 1$}
        \While{$i < \bfm[k]$}
            \State $\bfs = \bfs + P_{i*}$, where $P_{i*}$ is the $i$th dimension of all samples;
            \State i = i + 1;
        \EndWhile
        \State Sorting $\bfs$ to find a threshold $\bft[k]$ to let $\frac{\#(\bfs > \bft[k])}{n} > \bfv[k]$.
    \EndFor
\end{algorithmic}
\end{algorithm}

In the testing phase, two feature templates are matched from coarse to fine by Equation~(\ref{equ:dot_product}),
\ie from stage7 to stage1. If the similarity score is smaller than the threshold of current stage, the matching
process is interrupted. Because the probability of that the two feature templates coming from different subjects
is very high, most of the matching pairs will be rejected by the early stages of the cascade. Therefore, the
cascade can improve the matching speed significantly. The testing procedure is shown in the Alg.~\ref{alg:cascade_testing}.
Compared with ordinary linear scan way, the cascade only need two extra lines of code. This supplies a easy way to
improve the speed of existing large scale face recognition system.

\begin{algorithm}
\caption{Fast feature matching by the cascade classifier.}
\label{alg:cascade_testing}
\begin{algorithmic}[1]
\State {\bf Input}: Two normalized feature templates $\bfx'$ and $\bfy'$, feature size $d$,
    feature size of each stage $\bfm$, thresholds $\bft$, stage count $sn$.
\State {\bf Output}: Similarity $s$.
    \State Calculate the similarity in an incremental way:
    \State Let $s = 0$, $i = 0$;
    \For{$k = 0$ to $sn - 1$}
        \While{$i < \bfm[k]$}
            \State $s = s + \bfx'[i] \times \bfy'[i]$;
            \State i = i + 1;
        \EndWhile

            \color{red}
        \If{$s < \bft[k]$}
            \State break;
            \color{black}
        \EndIf

    \EndFor
\State return $s$.
\end{algorithmic}
\end{algorithm}

\section{Experiments}

\subsection{Database}

We use the FERET database to illustrate the advantages of our fast matching method following the standard
test protocols~\cite{Phillips-FERET-PAMI-00}. The results of the four popular experiments are reported:
fafb, fafc, fadup1 and fadup2.
In the experiments, fa is always used as gallery, and fb, fc, dup1 and dup2 are used as probe respectively.
The first two experiments fafb and fafc are already saturated by the most of state-of-the-art algorithms, but fadup1 and
fadup2 are still challenging due to the appearance variations caused by expression and aging.

To simulate large scale face recognition applications, we collect a large database, which contains $200,000$
face images with frontal pose, uniform illumination and good quality (These images can not be disclosed due to privacy issues).
We use the self-collected database to
enlarge the gallery of the FERET database to test the performance of the baseline method on the setting of
big data. With a big gallery, we can observe the variation of recognition rate and speed-up ratio of the cascade more easily.
Furthermore, we will analyze the relationship between the performance and feature size to supply a reference
for feature cutting in resource constrained applications.

\subsection{Baseline}

First, all face images in the training set, fa, fb, dup1 and dup2 are processed by a face detector and the facial landmarker.
We get 76 facial landmarks of each face image. Then, face images are normalize to a standard frame with eye distance $= 60$
pixels, and processed by DoG~\cite{Tan-TIP-2010} filter to alleviate the affection of illumination. Note that,
in the face normalization step, the 76 facial landmarks are transformed in the same way.

At the transformed 76 facial landmarks, $40\times76=3040$ Gabor features are extracted, and the dimension of Gabor feature
is reduced to 428 (The number of subjects in the training set is 429) by LDA. Finally, the similarity is evaluated by Cosine metric. The rank-1 recognition rate of the four experiments are shown in Table~\ref{tbl:baseline}. From that, we can see the baseline method is comparable with many state-of-the-art methods~\cite{Tan-TIP-2010, Xie-TIP-2010, Hussain-BMVC-2012}.

While the other
methods need to extract dense features from the face image, the proposed baseline only process the 76 sparse landmarks.
This will help to construct more efficient face recognition system in practical applications.
Furthermore, the baseline method has a potential advantage that is pose robustness. No matter what the pose of face is,
we always extract features at the 76 facial landmarks with fixed semantics.

\begin{table}
\caption{The baseline rank-1 recognition rate of the four experiments on the FERET database.}
\label{tbl:baseline}
\centering
\begin{tabular}{|c|c|c|c|c|}
  \hline
  Experiment & fafb & fafc & fadup1 & fadup2 \\
  \hline
  Tan \& Triggs~\cite{Tan-TIP-2010} & 98.0\% & 98.0\% & 90.0\% & 85.0\% \\
  \hline
  S-LGBP+LGXP~\cite{Xie-TIP-2010} & 99.0\% & 99.0\% & 94.0\% & 93.0\% \\
  \hline
  G-LQP~\cite{Hussain-BMVC-2012} & 99.9\% & 100\% & 93.2\% & 91.0\% \\
  \hline
  Proposed Baseline & 99.67\% & 100\% & 93.35\% & 92.74\% \\
  \hline
\end{tabular}
\end{table}

\subsection{Large Scale Gallery}

\begin{figure}
\centering
\includegraphics[width=0.45\textwidth]{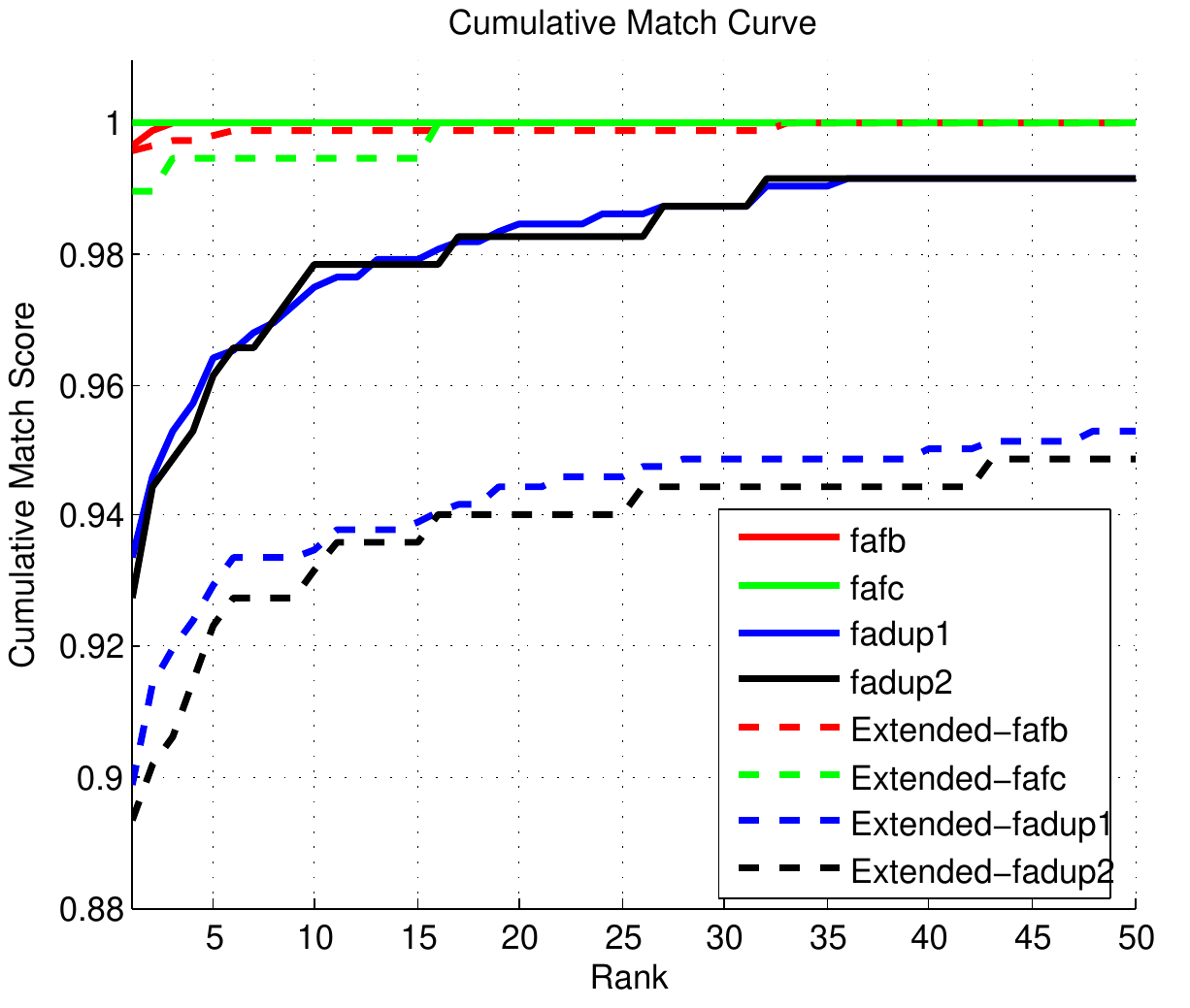}
\caption{The performance comparison of the baseline method before and after the gallery expansion, where ``Extended-''
denote the gallery fa is extended by the extra $200,000$ face images.}
\label{fig:large_scale_rr}
\end{figure}

To illustrate the performance in big data setting, we mix $200,000$ face images into the existing
gallery as ``Extended-fa'', and the probe sets remain unchanged. Like fa set, the Extended-fa is also processed
by the pipeline described in Section~\ref{sec:baseline}. Figure~\ref{fig:large_scale_rr} shows the impact
of the size of gallery to the recognition rate. We can see the last two experiments are more easily
affected by the size of gallery than the first two. However, when the Extended-fa contains $200,000 + 1196$ images,
the baseline still perform well. The rank-1 recognition rate of the last two experiments drop about $4\%$, \ie Extended-fadup1: from $93.35\%$ to $89.89\%$, Extended-fadup2: from $92.74\%$ to $89.32\%$.

In the following, the performance of the cascade and partial cascade are also evaluated on the large
extended gallery ``Extended-fa''.

\subsection{Cascade Evaluation}

The cascade is trained using the training set of the FERET and the settings described in
Section~\ref{sec:cascade}. Its performance is assessed in two aspects: recognition rate
loss, and computation speed up.

\begin{table*}
\caption{Evaluation of recognition rate and speed-up ratio of the cascade classifier on the extended FERET.}
\label{tbl:speed_up}
\centering
\begin{tabular}{|c|c|c|c|c|c|c|c|}
  \hline
   & Extended-fafb & Extended-fafc & Extended-fadup1 & Extended-fadup2 & Total Time & Time/Query & Speed-up Ratio\\
  \hline
  Normal & $99.58\%$ & $98.97\%$ & $89.89\%$ & $89.32\%$ &  &  &\\
  \cline{2-5}
         &  154894ms      & 25584ms     &  92977ms      & 30311ms       &   303766ms & 129.48ms & 1\\
  \hline
  Cascade & $99.58\%$   & $98.97\%$  & $89.89\%$    & $89.32\%$ &    &  & \\
  \cline{2-5}
        &   21060ms      & 3105ms     &  11981ms      & 4103ms       &   40249ms & 17.16ms & {\bf 7.55}\\
  \hline
\end{tabular}
\end{table*}

All experiments are conducted on the extended FERET database. The experimental platform is a normal PC with 2.4GHz CPU, 4G Memory. Table~\ref{tbl:speed_up} list the rank-1 recognition rate of normal and cascade based matching methods. It's amazing that the recognition rate of the cascade keeps as same as the original Cosine metric with zero performance loss. The $7th$ column of the table shows the average time to scan the gallery one time of all experiments, which indicate that the average matching speed is improved significantly by 7.55 times.

\subsection{Feature Length vs. Recognition Rate}

\begin{figure}
\centering
\includegraphics[width=0.45\textwidth]{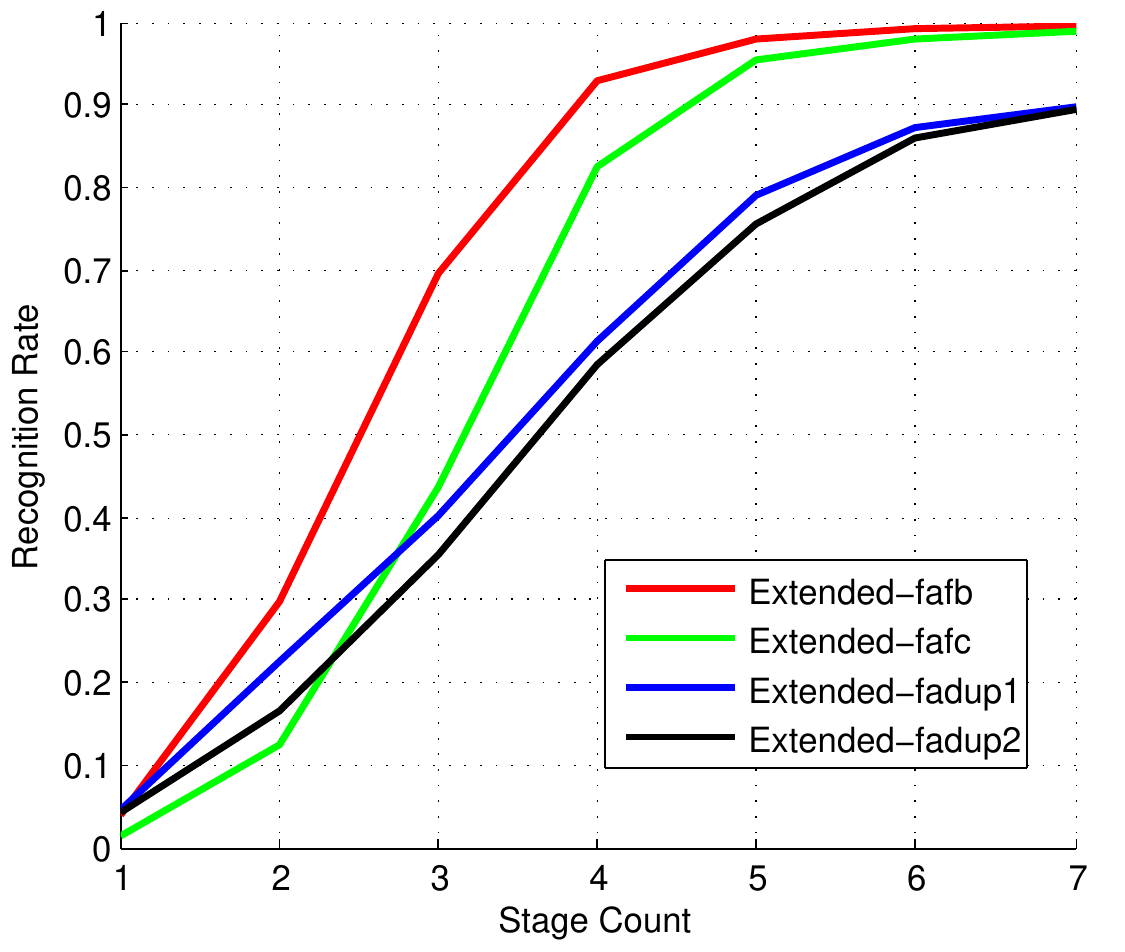}
\caption{The relationship between the recognition rate and feature length on the extended FERET database.}
\label{fig:RRvsFS}
\end{figure}

On one hand, the cascade provides a fast matching method for large scale face recognition systems. On the other hand,
it also provides a flexible structure for the system to cut off feature length to save the storage and
transmission bandwidth. Here, Figure~\ref{fig:RRvsFS} gives the relationship between the recognition rate and
feature length for reference.

We test the four experiments using partial cascade that means we just evaluate the similarity of two samples by the first $k$ stages in the cascade and neglect the last several stages. The x-axis in Figure~\ref{fig:RRvsFS} is the count of stage we used. From the figure, we can see the speed of improvement of recognition rate with respect to the stage count (or feature length)
is decrease. That means, as the length of feature increases, the recognition rate will become harder and
harder to improve. Therefore, we can cut off a large part at the end of the feature to get a
smaller feature template, while not sacrificing much recognition rate. For example, when the length of feature
template is cut from 428 to 214 dimension, \ie use 6 stages, the recognition rate of experiment ``fadup2'' just drop $3.42\%$.
This property supplies a large room for system-level optimization for some specific applications.

\section{Conclusions}

Computation speed, storage capacity, and transmission bandwidth are three key factors for large scale
face recognition systems. To improve the efficiency of system in terms of these factors, we propose
a cascade classifier and its learning algorithm. Through extensive experiments on the FERET and a large self-collected database,
we find the cascade can improve the face matching speed by 7.55 times with zero recognition rate loss.
The cascade also supplied a flexible structure for resource constrained applications, by which we could drop half of features just
with minor performance loss. Furthermore, the proposed baseline method (landmark based Gabor+LDA) is
comparable to many state-of-the-art methods on the FERET database, which will be studied further in the future
and applied in unconstrained face recognition problems, \eg LFW.

\section*{Acknowledgements}
The authors would like to acknowledge the following funding sources: the Chinese National
Natural Science Foundation Project \#61070146, the National Science and Technology Support
Program Project \#2009BAK43B26, the AuthenMetric R\&D Funds (2004-2012), and the TABULA
RASA project (http://www.tabularasa-euproject.org) under the Seventh Framework Programme
for research and technological development (FP7) of the European Union (EU), grant agreement
\#257289.

{\small
\bibliographystyle{ieee}
\bibliography{FastMatching}

\begin{thebibliography}{10}\itemsep=-1pt

\bibitem{face-com}
{face.com}.
\newblock http://www.face.com.

\bibitem{Ahonen-ECCV-04}
T.~Ahonen, A.~Hadid, and M.Pietikainen.
\newblock ``{Face} recognition with local binary patterns''.
\newblock In {\em {Proceedings of the European Conference on Computer Vision}},
  pages 469--481, Prague, Czech, 2004.

\bibitem{Belhumeur-97}
P.~Belhumeur, J.~Hespanha, and D.~Kriegman.
\newblock ``{Eigenfaces} vs. fisherfaces: recognition using class specific
  linear projection''.
\newblock {\em IEEE Trans. PAMI}, 19(7):711--720, 1997.

\bibitem{Bennett-MILCOM-2008}
B.~Bennett and C.~Dee.
\newblock ``{Scalable} video coding across heterogeneous networks''.
\newblock In {\em MILCOM, Military Communications Conference}, 2008.

\bibitem{Bourdev-CVPR-2005}
L.~Bourdev and J.~Brandt.
\newblock ``{Robust} object detection via soft cascade''.
\newblock In {\em Proceedings of IEEE Computer Society Conference on Computer
  Vision and Pattern Recognition}, pages 236--243, 2005.

\bibitem{Cootes-ASM-95}
T.~F. Cootes, C.~J. Taylor, D.~H. Cooper, and J.~Graham.
\newblock ``{Active} shape models: Their training and application''.
\newblock {\em CVGIP: Image Understanding}, 61:38--59, 1995.

\bibitem{Datar-SCG-2004}
M.~Datar, N.~Immorlica, P.~Indyk, and V.~S. Mirrokni.
\newblock ``{Locality}-sensitive hashing scheme based on p-stable
  distributions''.
\newblock In {\em Proceedings of the twentieth annual symposium on
  Computational geometry}, SCG '04, pages 253--262, New York, NY, USA, 2004.
  ACM.

\bibitem{friedman-an-algorithm-toms-77}
J.~H. Friedman, J.~L. Bentley, and R.~A. Finkel.
\newblock ``{An} algorithm for finding best matches in logarithmic expected
  time''.
\newblock {\em ACM Trans. Math. Softw.}, 3(3):209--226, September 1977.

\bibitem{Fukunaga-book-90}
K.~Fukunaga.
\newblock {\em Introduction to statistical pattern recognition}.
\newblock Academic Press, Boston, 2 edition, 1990.

\bibitem{Guo-ICCVW-2001}
G.-D. Guo and H.-J. Zhang.
\newblock ''{Boosting} for fast face recognition''.
\newblock In {\em Recognition, Analysis, and Tracking of Faces and Gestures in
  Real-Time Systems, 2001. Proceedings. IEEE ICCV Workshop on}, pages 96 --100,
  2001.

\bibitem{Huang-ICPR-2004}
C.~Huang, H.~Ai, B.~Wu, and S.~Lao.
\newblock ``{Boosting} nested cascade detector for multi-view face detection''.
\newblock {\em ICPR}, pages 415--418, 2004.

\bibitem{Lades-Gabor-93}
M.~Lades, J.~Vorbruggen, J.~Buhmann, J.~Lange, C.~von~der Malsburg, R.~P.
  Wurtz, and W.~Konen.
\newblock ``{Distortion} invariant object recognition in the dynamic link
  architecture''.
\newblock {\em IEEE Transactions on Computers}, 42:300--311, 1993.

\bibitem{Liu-ECCV-2008}
Y.~M. Lui and J.~R. Beveridge.
\newblock ``{Grassmann} registration manifolds for face recognition''.
\newblock In {\em ECCV}, pages 44--57, 2008.

\bibitem{Milborrow-PRASA-2010}
S.~Milborrow, J.~Morkel, and F.~Nicolls.
\newblock {The MUCT Landmarked Face Database}.
\newblock {\em Pattern Recognition Association of South Africa}, 2010.
\newblock http://www.milbo.org/muct.

\bibitem{Okada-Bochum-98}
K.~Okada, J.~Steffens, T.~Maurer, H.~Hong, E.~Elagin, H.~Neven, and C.~von~der
  Malsburg.
\newblock ``{The Bochum/USC Face Recognition System and How it Fared in the
  FERET Phase III Test}'', 1998.

\bibitem{Phillips-FERET-PAMI-00}
P.~J. Phillips, H.~Moon, S.~A. Rizvi, and P.~J. Rauss.
\newblock ``{The FERET} evaluation methodology for face-recognition
  algorithms''.
\newblock {\em IEEE Transactions on Pattern Analysis and Machine Intelligence},
  22(10):1090--1104, 2000.

\bibitem{Rother-ACM-TG-2004}
C.~Rother, V.~Kolmogorov, and A.~Blake.
\newblock ``{GrabCut}: interactive foreground extraction using iterated graph
  cuts''.
\newblock {\em ACM Trans. Graph.}, 23(3):309--314, Aug. 2004.

\bibitem{Saragih-IJCV-2011}
J.~Saragih, S.~Lucey, and J.~Cohn.
\newblock ``{Deformable} model fitting by regularized landmark mean-shift''.
\newblock {\em International Journal of Computer Vision}, 91:200--215, 2011.

\bibitem{Tan-TIP-2010}
X.~Tan and B.~Triggs.
\newblock ``{Enhanced} local texture feature sets for face recognition under
  difficult lighting conditions''.
\newblock {\em IEEE Transactions on Image Processing}, 19:1635--1650, June
  2010.

\bibitem{Hussain-BMVC-2012}
S.~ul~Hussain, Wheeler, T.~Napol¨¦on, and F.~Jurie.
\newblock ``{Face} recognition using local quantized patterns''.
\newblock In {\em Proc. British Machine Vision Conference}, volume~1, pages
  52--61, 2012.

\bibitem{Viola-IJCV-04}
P.~Viola and M.~Jones.
\newblock ``{Robust} real-time face detection''.
\newblock {\em International Journal of Computer Vision}, 57:137--154, 2004.

\bibitem{Wiskott-PAMI-97}
L.~Wiskott, J.~Fellous, N.~Kruger, and C.~v.~d. Malsburg.
\newblock ``{Face} recognition by elastic bunch graph matching''.
\newblock {\em IEEE Transactions on Pattern Analysis and Machine Intelligence},
  19(7):775--779, 1997.

\bibitem{Wu-PAMI-2011}
Z.~Wu, Q.~Ke, J.~Sun, and H.-Y. Shum.
\newblock ``{Scalable} face image retrieval with identity-based quantization
  and multireference reranking''.
\newblock {\em IEEE Trans. Pattern Anal. Mach. Intell.}, 33(10):1991--2001,
  Oct. 2011.

\bibitem{Xiao-ICCV-2003}
R.~Xiao, L.~Zhu, and H.-J. Zhang.
\newblock ``{Boosting} chain learning for object detection''.
\newblock In {\em Proceedings of the Ninth IEEE International Conference on
  Computer Vision}, ICCV '03, pages 709--715, Nice, France, 2003.

\bibitem{Xie-TIP-2010}
S.~Xie, S.~Shan, X.~Chen, and J.~Chen.
\newblock ``{Fusing} local patterns of gabor magnitude and phase for face
  recognition''.
\newblock {\em IEEE Transactions on Image Processing}, 19(5):1349--1361, 2010.

\bibitem{Yan-IJCB-2011}
J.~Yan, Z.~Lei, D.~Yi, and S.~Z. Li.
\newblock ``{Towards} incremental and large scale face recognition''.
\newblock In {\em International Joint Conference on Biometrics (IJCB)}, pages
  33--39, Washington, DC, USA, Oct. 11-13 2011.

\bibitem{Yi-IJCB-2011}
D.~Yi, Z.~Lei, and S.~Z. Li.
\newblock ``{A} robust eye localization method for low quality face images''.
\newblock In {\em International Joint Conference on Biometrics (IJCB)}, pages
  15--21, Washington, DC, USA, Oct. 11-13 2011.

\end{thebibliography}
}

\end{document}